\documentclass[10pt,twocolumn,letterpaper]{article}

\usepackage[]{cvpr}      

\definecolor{cvprblue}{rgb}{0.21,0.49,0.74}
\usepackage[pagebackref,breaklinks,colorlinks,allcolors=cvprblue]{hyperref}
\usepackage{xcolor}         
\usepackage{graphicx}
\usepackage{bm}
\usepackage{bbm}
\usepackage{colortbl}
\usepackage{pifont}
\usepackage{multicol}
\usepackage{multirow}
\usepackage{makecell}
\usepackage{wrapfig}
\usepackage{float}

\def\paperID{5562} 
\def\confName{CVPR}
\def\confYear{2025}

\title{From Obstacles to Resources: \\Semi-supervised Learning Faces Synthetic Data Contamination}

\author{
    Zerun Wang$^1$, Jiafeng Mao$^{1}$,Liuyu Xiang$^2$, Toshihiko Yamasaki$^1$\\
    $^1$The University of Tokyo;\\
    $^2$Beijing University of Posts and Telecommunications;\\
    {\tt \small \{ze\_wang,yamasaki\}@cvm.t.u-tokyo.ac.jp}\\ 
    {\tt \small mao@hal.t.u-tokyo.ac.jp},
    {\tt \small xiangly@bupt.edu.cn}\\
}

\begin{document}
\maketitle



\begin{abstract}
Semi-supervised learning (SSL) can improve model performance by leveraging unlabeled images, which can be collected from public image sources with low costs. In recent years, synthetic images have become increasingly common in public image sources due to rapid advances in generative models. Therefore, it is becoming inevitable to include existing synthetic images in the unlabeled data for SSL. How this kind of contamination will affect SSL remains unexplored. In this paper, we introduce a new task, Real-Synthetic Hybrid SSL (RS-SSL), to investigate the impact of unlabeled data contaminated by synthetic images for SSL. First, we set up a new RS-SSL benchmark to evaluate current SSL methods and found they struggled to improve by unlabeled synthetic images, sometimes even negatively affected. To this end, we propose RSMatch, a novel SSL method specifically designed to handle the challenges of RS-SSL. RSMatch effectively identifies unlabeled synthetic data and further utilizes them for improvement. Extensive experimental results show that RSMatch can transfer synthetic unlabeled data from `obstacles' to `resources.' The effectiveness is further verified through ablation studies and visualization.
\end{abstract}

    
\section{Introduction}
\label{sec:intro}

Semi-supervised learning (SSL) methods~\cite{sohn2020fixmatch,zhang2021flexmatch,chen2023softmatch} require a small number of labeled data for supervision and leverage extensive unlabeled data to enhance model performance, thereby reducing the need for extensive data labeling. 
A major origin of unlabeled images is the publicly available image sources, \eg, the Internet.



\begin{figure}
  \centering
  \includegraphics[width=0.99\linewidth]{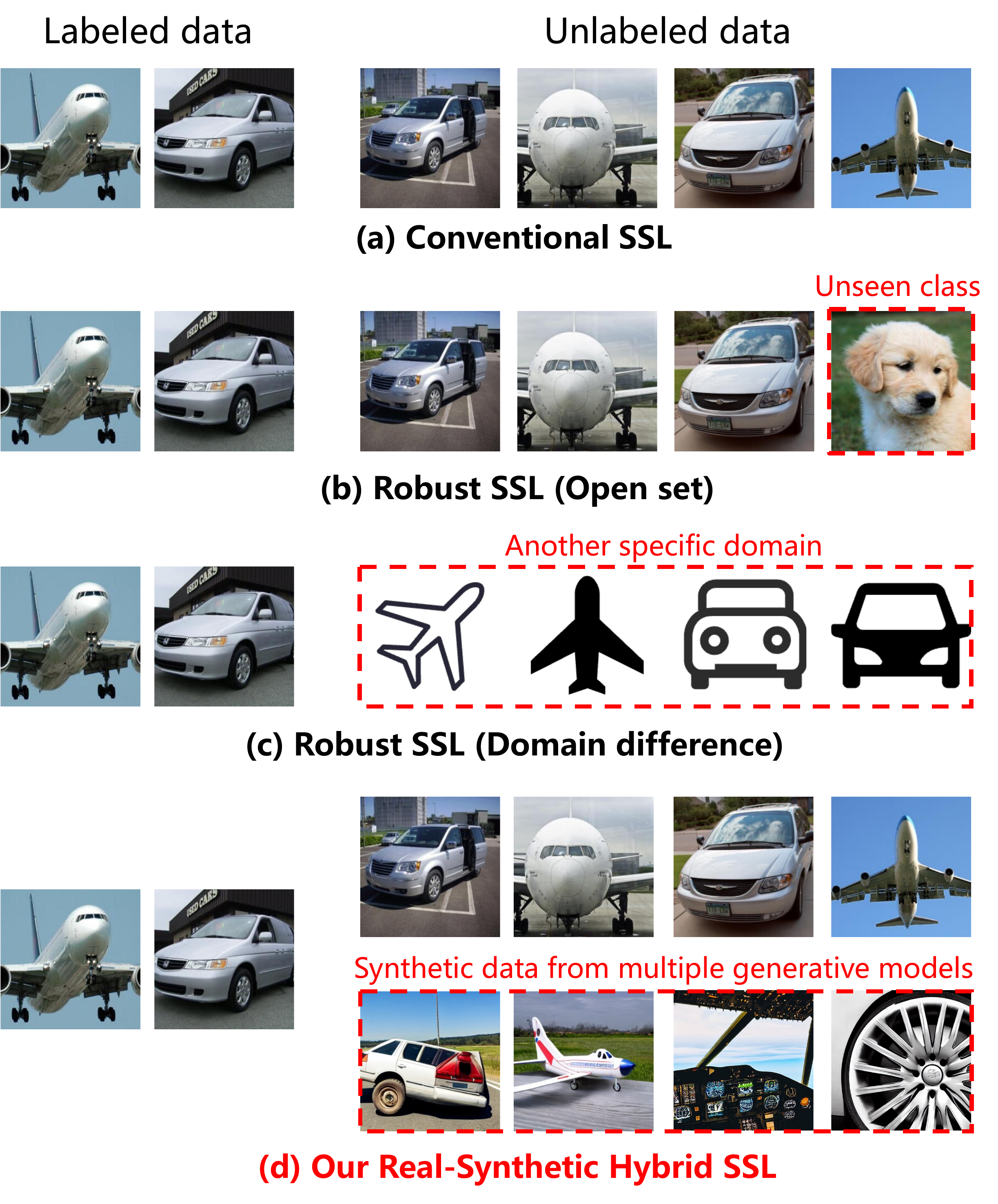}
  \caption{The task setting of Real-Synthetic Hybrid SSL (RS-SSL) compared with previous SSL tasks. The unlabeled dataset for SSL is contaminated by existing synthetic images uploaded to public image sources. Note that there is no annotation for real or synthetic in unlabeled data.}
  \label{fig1}
\end{figure}

Various SSL methods are proposed to improve the effectiveness of learning from both labeled and unlabeled data. However, a new problem has arisen in recent years: A large number of powerful consumer-grade generative models~\cite{ho2020denoising,nichol2022glide} is emerging rapidly, allowing users to enjoy the convenience of synthesizing and sharing AI-generated images. Thus, an increasing amount of synthetic data is being uploaded to public image sources. As a result, the unlabeled data we collect from public image sources will be inevitably contaminated by synthetic data, as shown in Fig.~\ref{fig1}. 

The problem introduced by synthetic data contamination in the unlabeled data for SSL has not been investigated: As shown in Fig.~\ref{fig1}, although some robust SSL methods~\cite{saito2021openmatch,li2023iomatch,huang2021universal,jia2023bidirectional} focused on different domains of the labeled and unlabeled data, all the unlabeled data are from another specific domain with significant different visual appearance in their setting. However, in our setting, the unlabeled data includes both real images from the target domain and synthetic data from various generative models. And these synthetic images are sometimes hard to identify, even for humans. Meanwhile, some works~\cite{yamaguchi2024generative,yamaguchi2024regularizing} focused on designing generative models for SSL, not investigating the impact of existing synthetic images. Thus, their task is different from ours. 




Given this problem unexplored, we first investigate: \textbf{Are SSL methods ready for unlabeled data contaminated by synthetic images?} To answer this question, we set up a new task named Real-Synthetic Hybrid SSL (RS-SSL), which means that a portion of existing synthetic images from public image sources is mixed into the unlabeled data, as illustrated in Fig.~\ref{fig1}. For this task, we established a new RS-SSL benchmark with commonly used SSL datasets and mainstream generative models. Then, we evaluated the performance of some representative SSL methods on our benchmark. Unfortunately, our finding is that current SSL methods, even those robust SSL methods, cannot utilize existing synthetic data efficiently: The contamination caused by existing synthetic images provides no significant improvement and sometimes has a negative impact. 

\textbf{Can we alleviate this problem and better utilize existing synthetic images for SSL?} We analyzed the negative impact caused by the synthetic images and found the potential utilization value of these data. We further propose a new SSL method, RSMatch, for utilizing these synthetic data. RSMatch deals with the unlabeled data in two steps: 

\noindent1) \textit{Synthetic data identification}: RSMatch can distinguish the synthetic images in the unlabeled data. This is challenging since we have no labeled synthetic images, which makes it impossible to train a synthetic image identification model using standard supervision. RSMatch achieves this by dynamically mining reliable synthetic data for training a deepfake detector. 

\noindent2) \textit{SSL with real and synthetic data}: With the success of identification, RSMatch can eliminate the negative impact and further utilize the synthetic images to improve SSL performance by introducing an extra dummy head for dealing with synthetic data.

Our contributions can be summarized as follows:
\begin{itemize}
    \item We introduce RS-SSL, a new practical and challenging SSL task considering the impact of synthetic data contamination.
    \item We observed that recent SSL methods struggle to utilize unlabeled synthetic data on our RS-SSL benchmark and analyzed the issue caused by these data.  
    \item We propose RSMatch, a new SSL method that can identify and utilize the synthetic unlabeled data to improve SSL performance.
\end{itemize}
\section{Related work}
\subsection{Semi-supervised learning}
Semi-supervised learning (SSL) aims to utilize unlabeled data to improve the model supervised by limited labeled data, thus reducing the cost of labeling additional data. Typically, SSL methods apply consistency regularization~\cite{bachman2014learning} for self-training to utilize unlabeled data. FixMatch~\cite{sohn2020fixmatch} is one of the most influential SSL methods recently, as it effectively integrates some techniques from earlier works~\cite{berthelot2019mixmatch,xie2020unsupervised,rasmus2015semi,berthelot2019remixmatch}, becoming a foundational framework for subsequent works. The following works improved the effectiveness of SSL in various aspects, such as class-wise dynamical thresholds~\cite{zhang2021flexmatch, wang2023freematch}, contrastive learning strategies~\cite{li2021comatch, zheng2022simmatch}. These methods have also propelled the development of SSL for downstream tasks~\cite{sohn2020simple,papandreou2015weakly}. Meanwhile, there are also SSL methods, named robust SSL, considering the potential issues in unlabeled data, such as class imbalance~\cite{yu2022inpl,guo2022class}, domain difference~\cite{huang2021universal,jia2023bidirectional}, and open-set problems~\cite{li2023iomatch, saito2021openmatch}. However, the issue caused by synthetic data contamination remains uninvestigated. And we experimentally found that these robust SSL methods cannot tackle this problem.

As far as we know, we are the first to tackle the problem of SSL with synthetic data contamination. While some works such as MP-SSL~\cite{yamaguchi2024generative} considered a problem named generative SSL, their setting is to guide the generative model to generate better data for training. Thus, it is different from ours, as we aim to mitigate the impact caused by already existing synthetic data in the unlabeled data.

\begin{figure*}
  \centering
  \includegraphics[width=0.99\linewidth]{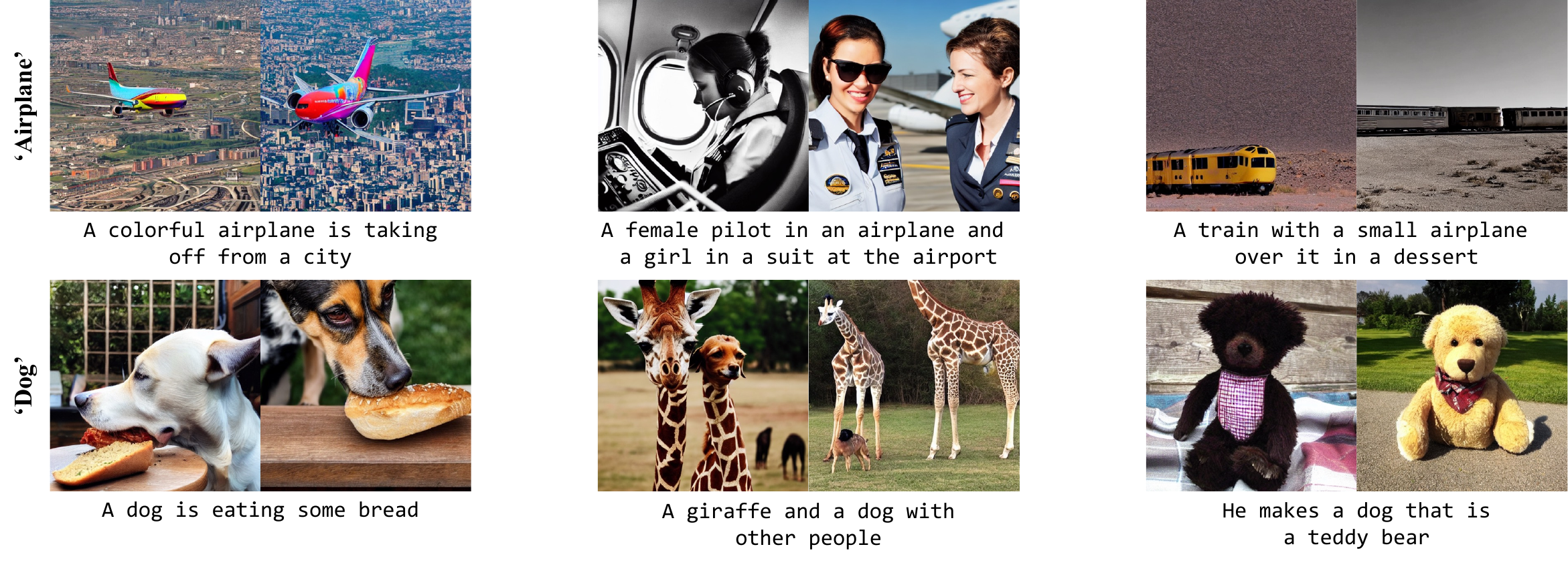}
  \caption{Examples of generation results for class 'airplane' and 'dog' in CIFAR-10~\cite{krizhevsky2009learning} with the SD14 model~\cite{rombach2022high}. The prompts under the images are automatically generated by the T5 model~\cite{raffel2020exploring}. Note that besides images matching the target class, there are also some images with semantic bias and even some entirely unrelated images. This makes our RS-SSL benchmark closer to the practical scenario.}
  \label{fig2}
\end{figure*}

\subsection{Training with synthetic data}
Utilizing synthetic data from training has never ceased, as this kind of data is easy to obtain. 
With the rapid emergence of powerful generative models, How synthetic images impact training has become an unavoidable problem. Early works~\cite{besnier2020dataset,zhang2021datasetgan,jahanian2021generative} explored training with images generated by GAN-based models. Recent works focus on generated data from more powerful Diffusion Models~\cite{sohl2015deep,ho2020denoising}. Hataya \etal~\cite{hataya2023will} evaluated the effect of generated data on several supervised vision tasks and found that the generated data can have negative effects. He \etal~\cite{he2022synthetic} found the generated data from the GLIDE model~\cite{nichol2022glide} is helpful for zero-shot and few-shot classification on the pre-trained CLIP model~\cite{radford2021learning}. These works, however, only focus on investigating the impact of labeled synthetic data on supervision tasks where the prompt is available and serves as the classification label. In contrast, we consider another practical scenario for SSL, in which the collected real and synthetic images are all unlabeled and mixed together.

\subsection{Text-to-image diffusion model}
Diffusion models~\cite{sohl2015deep,ho2020denoising} learn to reverse the forward diffusion process to recover the original data from the added noise. Due to the high generation quality, diffusion models have become the mainstream generative models in recent years. Meanwhile, diffusion models can be guided by multimodal conditions to synthesize images~\cite{meng2021sdedit,zhang2023adding}. Text-to-image generation takes the natural language description as the condition and is supported by many diffusion model-based works such as GLIDE~\cite{nichol2022glide}, stable diffusion~\cite{rombach2022high} with its variants~\cite{podell2023sdxl,lin2024sdxl}, Imagen~\cite{saharia2022photorealistic}, \etc. These works made high-quality generative models accessible to ordinary users on consumer-grade devices, leading to a large number of synthetic images being uploaded to public image sources. We use a combination of these models to construct the RS-SSL benchmark closer to real-world scenarios.
\section{Are SSL methods ready for RS-SSL?}
\label{sec3}

\subsection{RS-SSL benchmark}
We build a new benchmark for evaluating SSL methods. We first analyze how existing synthetic data will be mixed in the unlabeled data during data collection. Then, we apply commonly used generative models and SSL datasets to construct the RS-SSL benchmark.

\noindent\textbf{Analysis on data collection.} When collecting unlabeled data from public image sources, one major way is using the class names of the target task as keywords for searching. Therefore, it is likely to get the public synthetic images attached by text prompts related to the target classes. Meanwhile, considering the vast number of generative AI users, the text prompts will be highly diverse, and the images will come from different generative models. Therefore, we propose to construct the RS-SSL benchmark as follows.

\noindent\textbf{Prompts.} To simulate the practical data collection scenario, we use diverse text prompts related to the classes of the target task for generation. Inspired by He \etal~\cite{he2022synthetic}, we leverage an off-the-shelf word-to-sentence model T5~\cite{raffel2020exploring} to get diverse description texts. Specifically, we feed the T5 model with class names from SSL datasets and generate various descriptions. For example, the generated description for the class name \texttt{cat} could be \texttt{A beautiful white cat laying on a rock}.

\noindent\textbf{Generative models.} While previous works~\cite{he2022safe,hataya2023will} focusing on the impact of synthetic data typically use one specific generative model, we consider that different users will use different types of generative models. Thus, we apply three different generative models, GLIDE~\cite{nichol2022glide}, Stable Diffusion v1.4 (SD14)~\cite{rombach2022high}, and SDXL-Lightening~\cite{lin2024sdxl}, to represent the commonly used generative models published at different times. We apply these models with the above-mentioned text prompts to get synthetic images.

\noindent\textbf{Benchmark construction.} 
We adopt commonly used SSL datasets as target tasks and construct the RS-SSL benchmark by adding additional synthetic images into the unlabeled data. Specifically, for each class in one selected SSL dataset, we use the T5 model to get $M$ text prompts and randomly use the prompts to generate $\alpha N$ images with our selected generative models. Each model generates the same number of images. Here, $N$ is the number of original unlabeled images, and we introduce \textit{synthetic ratio}, denoted as $\alpha$, to control the ratio of synthetic images. 
We keep the class-wise balance during generation. Fig.~\ref{fig2} demonstrates some synthetic images in our RS-SSL benchmarks. 



\subsection{Evaluating existing SSL methods.}

We evaluate three representative SSL methods~\cite{sohn2020fixmatch,zhang2021flexmatch,chen2023softmatch} and three robust SSL methods~\cite{saito2021openmatch,li2023iomatch,huang2021universal} with the constructed RS-SSL benchmark. To explore the impact of synthetic data, we mainly compare the SSL results of each method on 1) the entire RS-SSL benchmark and 2) the RS-SSL benchmark without the synthetic part (\ie, only the real unlabeled data). The results in Tables~\ref{tab1} and ~\ref{tab2} show that \textbf{existing synthetic images in unlabeled data cannot provide significant improvement and even have negative effects at times}. Meanwhile, the robust SSL methods originally designed for open-set and domain adaption problems also failed on this task.

\begin{figure}
  \centering
  \includegraphics[width=0.8\linewidth]{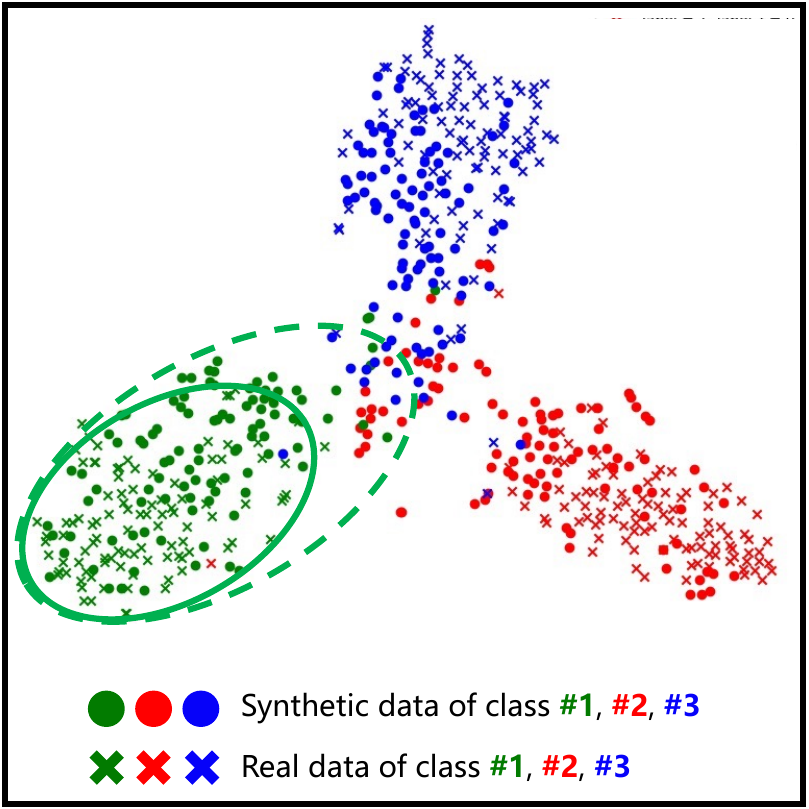}
  \caption{T-SNE~\cite{van2008visualizing} visualization of real and synthetic unlabeled samples to show the distribution bias. The samples are from three random classes in our benchmark with the CIFAR-10 dataset and an additional 50\% synthetic images in the unlabeled data (\ie, $\alpha=0.5$). The model is trained with FixMatch~\cite{sohn2020fixmatch}.}
  \label{fig3}
\end{figure}

\subsection{Analysis of the impact of synthetic data.} 

The failure of existing SSL methods prompts us to analyze the issue caused by the introduction of synthetic data. While previous research~\cite{hataya2023will,sariyildiz2023fake} has observed a bag of visual issues with synthetic data, \eg, unnatural details, semantic errors, \etc, we further go through the viewpoint of the model (\ie, the feature distribution) to analyze the impact of synthetic data for SSL. Fig.~\ref{fig3} is the T-SNE~\cite{van2008visualizing} feature distribution visualization of unlabeled images from three randomly selected classes from CIFAR-10. The feature is extracted by a learned model using the FixMatch~\cite{sohn2020fixmatch} SSL method. Our observation can be summarized as follows:

\noindent\textbf{The negative impact of synthetic data.} There exists a distribution bias between real and synthetic unlabeled data for each class. This is likely caused by style differences, semantic bias (\eg, a whole car in real data vs. a wheel in synthetic data), and noisy samples. Those biased synthetic samples break the boundaries between different classes, thus leading to a negative impact on SSL.

\noindent\textbf{The potential value of synthetic data.} Despite the distribution bias, we also observe that many synthetic samples form a good cluster and are close to the real data cluster of the related class. Meanwhile, from the visualization of synthetic samples, it is evident that some images effectively provide new content related to the target class, which can enrich the knowledge learned by the model. This suggests the potential learning value of the synthetic images. 


\section{RSMatch}

\begin{figure*}
\centering
\includegraphics[width=0.99\linewidth]{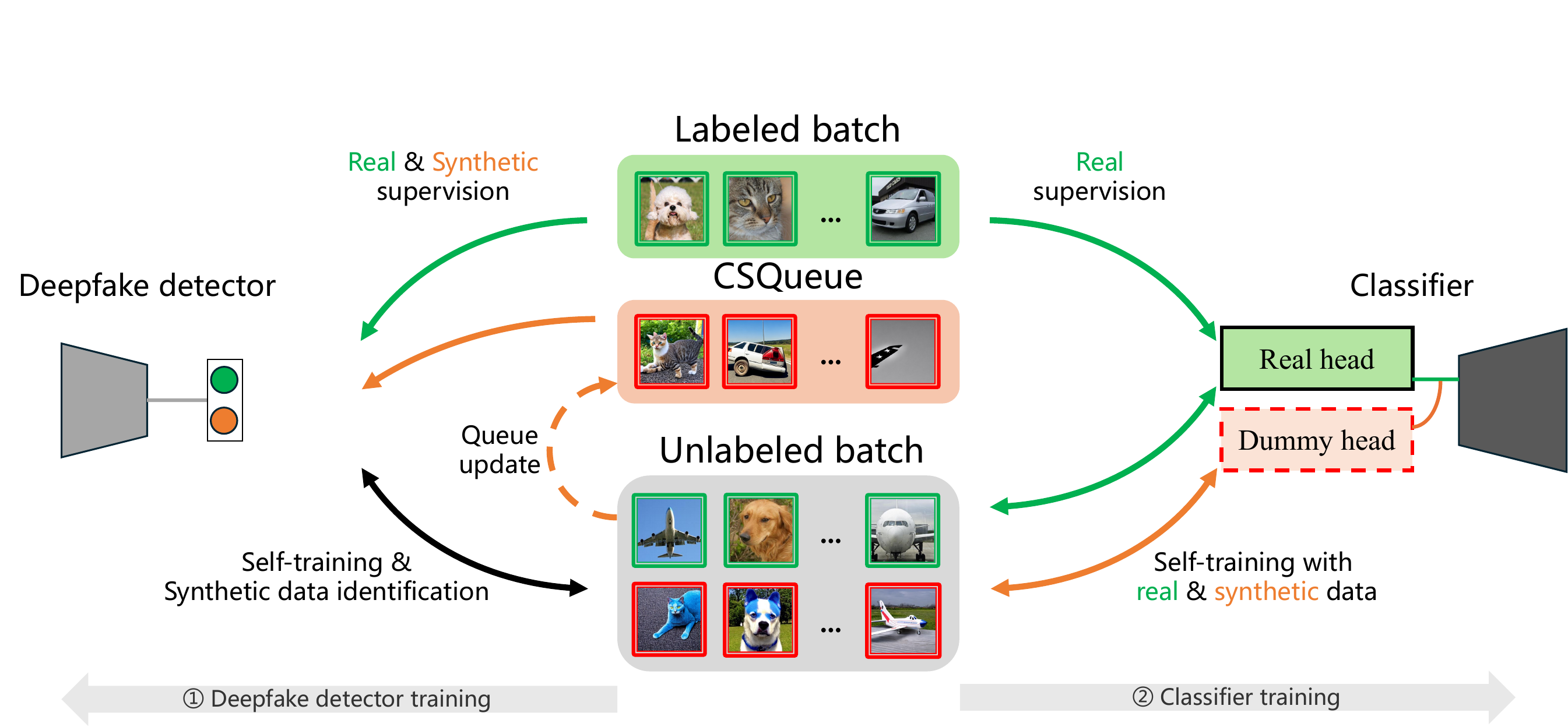}
\caption{The RSMatch framework. \textbf{Left}: A lightweight deepfake detector is trained to identify real and synthetic images in the unlabeled data batch. The labeled synthetic data for supervision comes from the CSQueue, which is proposed for mining and storing synthetic data from unlabeled data. \textbf{Right}: The classifier with the real and dummy head for self-training on the identified real and synthetic unlabeled data separately. The two networks are trained simultaneously.}
\label{fig4}
\end{figure*}

\begin{figure}
\centering
\includegraphics[width=0.99\linewidth]{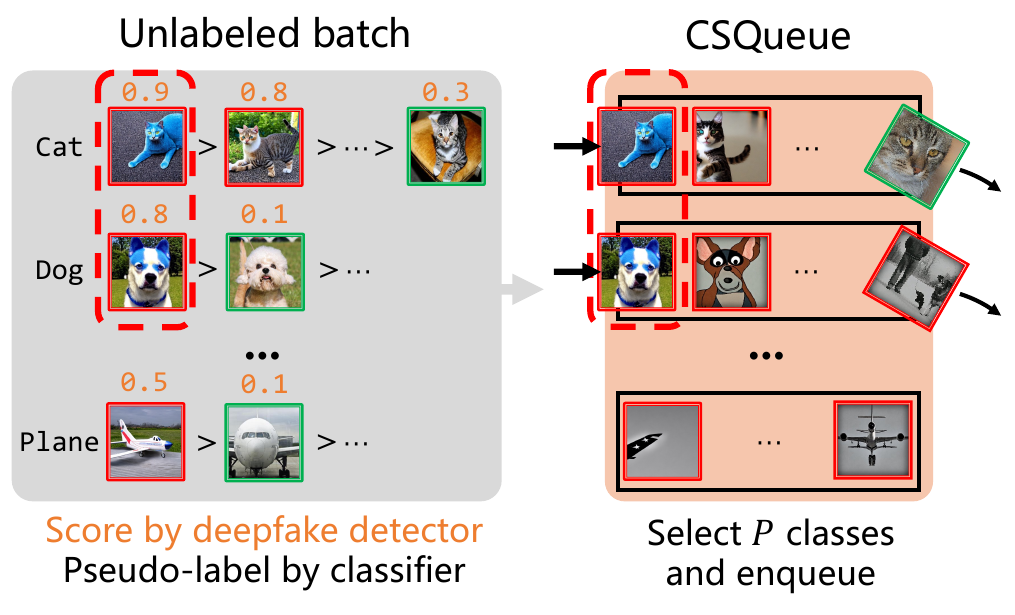}
\caption{Updating strategy of the CSQueue. The unlabeled data are pseudo-labeled by the classifier and then sorted by the confidence score from the deepfake detector. We select $P$ classes and push the top-$Q$ images for each sub-queue in one iteration.}
\label{fig5}
\end{figure}


The evaluation results confirm the necessity of introducing and solving the RS-SSL task. Based on our findings, we propose a new SSL method, RSMatch, to transfer synthetic data \textit{from obstacles to resources.}

\subsection{Overview of RSMatch}



As illustrated in Fig.~\ref{fig4}, RSMatch involves two steps in each training iteration. 

\noindent\textit{1) Synthetic data identification}: A lightweight deepfake detector, which is a binary classifier for real and synthetic, is trained to identify the synthetic unlabeled images. Since there is no labeled synthetic data, we propose the class-wise synthetic data queue (CSQueue) to dynamically mine reliable synthetic images from the unlabeled data for supervision, as detailed in Sec.~\ref{sec4_2}. 

\noindent\textit{2) SSL with both real and synthetic data}: The proposed real and dummy head structure is used for training with real and synthetic images separately to improve the SSL performance, as detailed in Sec.~\ref{sec4_3}.

RSMatch is orthogonal to the basic SSL methods. Thus, we can apply different methods as the basic framework for supervision and self-training strategies. Here, we use FixMatch~\cite{sohn2020fixmatch} as the basic SSL framework for introduction in the following sections.

\subsection{Synthetic data identification}
\label{sec4_2}

Training a deepfake detection model with real and synthetic labels is a mature technology~\cite{yan2024deepfakebench}. However, in the RS-SSL task, there is no labeled synthetic data. Thus, we propose to mine and store reliable synthetic images as new labeled data for supervision. This is achieved by our CSQueue and the corresponding updating and training strategy.

\noindent\textbf{CSQueue structure.}
CSQueue $\mathcal{D}^q$ contains $K$ first-in, first-out sub-queues. $K$ is the number of classes of the target task. The $i$-th sub-queue $\mathcal{D}^q_i$ is used to store the mined synthetic images $\{\bm{x}_k\}_{k=1}^{N_{q}}$ corresponding to the $i$-{th} class. $N_{q}$ is the queue size.

We propose a class-wise queue structure because we found a single queue will be dominated by synthetic data from a certain few classes. Thus, the model can only identify synthetic data from these classes. We believe this is because synthetic data from some classes are more easily distinguished than those from other classes.

\noindent\textbf{Updating strategy.} For each training iteration, we randomly select $P$ classes (\ie, $P$ sub-queues, $P \le K$) for updating. As shown in Fig.~\ref{fig5}, for each selected class, we first obtain the unlabeled images belonging to this class regarding the pseudo-labels from the classifier's real head $\mathcal{C}^{r}$.  Then, we sort the images based on $p^{\prime}_{s}$ from the deepfake detector, which denotes the confidence score to be synthetic data:
\begin{equation}
    p^{\prime}_{s} = \frac{e^{z^{\prime}_s}}{e^{z^{\prime}_r}+e^{z^{\prime}_s}},
\end{equation}
where $[z^{\prime}_r, z^{\prime}_s]$ are the logits output on `real' and `synthetic' respectively. Finally, we push the top-$Q$ images with the highest $p^{\prime}_{s}$ into the corresponding sub-queue. Meanwhile, the oldest samples will be popped if the length exceeds $N_{q}$. This first-in-first-out strategy can prevent noisy images from remaining for a long time.

\noindent\textbf{Training deepfake detector.} The images stored in CSQueue serve as newly labeled synthetic data. In each iteration, we sample a batch of real data $\mathcal{R}$ from the labeled dataset $\mathcal{D}^l$ and a batch of synthetic data $\mathcal{S}$ from CSQueue $\mathcal{D}^q$ with the same batch size $B$ to train the deepfake detector:
\begin{equation}
    \mathcal{L}^{\prime}_{s} = \frac{1}{2B}\sum_{i=1}^{2B}H(y^{\prime}_i, \mathcal{C}^{\prime}(\mathcal{F}^{\prime}(\bm{x}_i))),
\label{eq2}
\end{equation}
where $\bm{x}\in\mathcal{R}\cup\mathcal{S}$ is the input image. $y^{\prime}\in\{0, 1\}$ denotes the real and synthetic labels, respectively, $H$ denotes the cross-entropy loss function. $\mathcal{C}^{\prime}$, $\mathcal{F}^{\prime}$ denotes the binary head and encoder of the deepfake detector.

Meanwhile, the provided unlabeled batch from the SSL task enables us to further improve the deepfake detector by self-training following the basic SSL framework:
\begin{equation}
    \mathcal{L}^{\prime}_{u} = \frac{1}{\mu B}\sum_{i=1}^{\mu B}\mathbbm{1}(\max(q^{\prime}_i)>\tau)H(\hat{q}^{\prime}_i, \mathcal{C}^{\prime}(\mathcal{F}^{\prime}(\bm{u}_i))), 
\label{eq3}
\end{equation}
where $\bm{u}\in\mathcal{D}^u$ denotes the unlabeled images. We use a batch size of $\mu B$, where $\mu$ controls the ratio of labeled and unlabeled batches.  $q_i^{\prime}=\text{softmax}(\mathcal{C}^{\prime}(\mathcal{F}^{\prime}(\bm{u}_i))$ denotes the confidence score of the deepfake detector. $\hat{q}_i^{\prime}=\text{argmax}(q_i^{\prime})$ is the binary pesudo-label. $\mathbbm{1}(\max(q^{\prime}_i)>\tau)$ is the indicator for selecting samples with confidence scores over threshold $\tau$. This thresholding process follows FixMatch~\cite{sohn2020fixmatch}. Different thresholding methods will be applied regarding the basic SSL framework.

Experimental results prove that our method can mine reliable synthetic images from unlabeled data and achieve over 90\% identification accuracy. We also demonstrate the necessity of the class-wise design through ablation studies.

\subsection{SSL with both real and synthetic data}
\label{sec4_3}
The success of identifying synthetic images in unlabeled data enables us to further consider how to utilize them. With our analysis above, 1) The distribution bias between the synthetic and real images will corrupt the learned prototype for the target classes. 2) However, we believe that those well-clustered synthetic images also contain valuable content related to their corresponding class. Thus, treating them properly might enable the encoder to extract more comprehensive features.

Based on this idea, we propose to use identified synthetic images to train the encoder without affecting the original classification head since the parameters of the head are related to the learned class-wise prototype. To achieve this, we introduce an additional head, \ie, the dummy head, to the classifier for dealing with synthetic data. Thus, the original head, \ie, the real head, can be protected from synthetic data. Meanwhile, the gradients from synthetic data can also be backpropagated to the encoder.

\noindent\textbf{Training with real data.} The original real head of the classifier uses labeled real data for supervision and unlabeled real data for self-training following the basic SSL framework~\cite{sohn2020fixmatch}:
\begin{equation}
    \mathcal{L}^r_{s} = \frac{1}{B}\sum_{i=1}^{B}H(y_i, \mathcal{C}^{r}(\mathcal{F}(\bm{x}_i))),
\end{equation}
\begin{equation}
\resizebox{\linewidth}{!}{$
    \mathcal{L}^r_{u} = \frac{1}{\mu B}\sum_{i=1}^{\mu B}\mathbbm{1}(\max(q_i)>\tau)\mathbbm{1}(\hat{q}^{\prime}_i = 0)H(\hat{q}_i, \mathcal{C}^{r}(\mathcal{F}(\bm{u}_i))),$
}
\label{eq6}
\end{equation}
where $\bm{x} \in \mathcal{D}^l$, $\bm{u} \in \mathcal{D}^u$ denotes the labeled and unlabeled data, respectively. $\mathcal{C}^r$, $\mathcal{F}$ denotes the real head and encoder of the classifer. $q_i=\text{softmax}(\mathcal{C}^{r}(\mathcal{F}(\bm{u}_i))$ denotes the confidence score from the real head. $\hat{q}_i=\text{argmax}(q_i)$ is the K-class pseudo-label. $y_i$ denotes the K-class annotation of labeled data. $\mathbbm{1}(\hat{q}^{\prime}_i=0)$ denotes filtering out the synthetic images by the deepfake detector. 

\noindent\textbf{Training with synthetic data.} The identified synthetic images are further utilized to train the dummy head $\mathcal{C}^{d}$:
\begin{equation}
\resizebox{\linewidth}{!}{$
    \mathcal{L}^s_{u} = \frac{1}{\mu B}\sum_{i=1}^{\mu B}\mathbbm{1}(\max(q_i)>\tau)\mathbbm{1}(\hat{q}^{\prime}_i =  1)H(\hat{q}_i, \mathcal{C}^{d}(\mathcal{F}(\bm{u}_i))),
$}
\end{equation}
where $\mathbbm{1}(\hat{q}^{\prime}_i=0)$ is for only using the synthetic images. $\hat{q}_i$ is also the pseudo-label from the real head $\mathcal{C}^{r}$, as mentioned in Eq.~\ref{eq6}. The dummy head will not be used for pseudo-labeling.

Finally, RSMatch is trained with all the losses mentioned above:
\begin{equation}
    \mathcal{L} = \underbrace{\mathcal{L}^{\prime}_{s} + \mathcal{L}^{\prime}_{u}}_{\text{for deepfake detector}} + \underbrace{\mathcal{L}^r_{s} + \lambda(\mathcal{L}^r_{u} + \mathcal{L}^s_{u})}_{\text{for classifier}},
\end{equation}
where $\mathcal{L}^{\prime}_{s}$, $\mathcal{L}^{\prime}_{u}$ are applied to the deepfake detector. $\mathcal{L}^r_{s}$, $\mathcal{L}^r_{u}$, and $\mathcal{L}^s_{u}$ are applied to the classifier. $\lambda$ is the weight of unsupervised loss introduced by the basic SSL framework~\cite{sohn2020fixmatch}. 

\noindent\textbf{Inference.} RSMatch needs only the real head and the encoder from the classifier for inference. The deepfake detector, CSQueue, and dummy head are only used for training and can be easily removed during inference. Therefore, compared with basic SSL methods, RSMatch does not introduce any additional inference computational cost. 

\section{Experiments}

\subsection{Experimental setup}
\textbf{Benchmark and metrics.} We construct the RS-SSL benchmark following Sec.~\ref{sec3}. For the SSL datasets, we choose the commonly used CIFAR-10~\cite{krizhevsky2009learning}, CIFAR-100~\cite{krizhevsky2009learning}, and large-scale TinyImageNet~\cite{deng2009imagenet}, ImageNet~\cite{deng2009imagenet}.
We use various labeled data numbers and synthetic ratios $\alpha$ for experiments. We consider the standard classification accuracy for evaluation.

\noindent\textbf{Implementation details.} We generate $M=200$ prompts for each class in one dataset to construct the RS-SSL benchmark. We apply RSMatch on three representative SSL methods: FixMatch~\cite{sohn2020fixmatch}, FlexMatch~\cite{zhang2021flexmatch}, and SoftMatch~\cite{chen2023softmatch}. We also follow their settings on the SSL hyperparameters. For the classifier, we use WideResNet28-2~\cite{zagoruyko2016wide} for all the datasets except ImageNet, where we use ResNet-18~\cite{he2016deep}. For the deepfake detector, we apply the same structure but half the channels in each medium layer. The new hyperparameters introduced by RSMatch are the sub-queue size $N_q$, selected sub-queue number $P$, and enqueue number per iteration $Q$. We set $N_q=8$, $P=10$, and $Q=1$. The ablation of these parameters and other detailed settings are reported in the supplementary material. For the main results, we ran each experiment three times and reported the mean result with standard deviation. We use the USB~\cite{wang2022usb} SSL codebase.

\begin{table*}[t]
\centering
\begin{center}
\resizebox{\linewidth}{!}{
\begin{tabular}{l|ccc|ccc}
\toprule
Dataset & \multicolumn{3}{c|}{CIFAR-10} & \multicolumn{3}{c}{CIFAR-100}  \\
\midrule
Synthetic ratio ($\alpha$) & 0.3 & 0.5 & 1.0 & 0.3 & 0.5 & 1.0 \\
\midrule
FixMatch &  95.12\scalebox{0.8}{$\pm$0.3} \textcolor{teal}{(+0.16)} & 94.80\scalebox{0.8}{$\pm$0.2} \textcolor{red}{(-0.17)} & 92.86\scalebox{0.8}{$\pm$0.2} \textcolor{red}{(-1.89)} & 64.26\scalebox{0.8}{$\pm$0.2} \textcolor{red}{(-1.10)} & 64.14\scalebox{0.8}{$\pm$0.0} \textcolor{red}{(-0.38)} & 63.55\scalebox{0.8}{$\pm$0.3} \textcolor{red}{(-0.56)}\\
FixMatch (real only) &  94.96\scalebox{0.8}{$\pm$0.1} & 94.97\scalebox{0.8}{$\pm$0.1} & 94.75\scalebox{0.8}{$\pm$0.3} & 65.36\scalebox{0.8}{$\pm$0.1} & 64.52\scalebox{0.8}{$\pm$0.1} & 64.11\scalebox{0.8}{$\pm$0.2}\\
\textbf{Ours-FixMatch} &  \textbf{95.31\scalebox{0.8}{$\pm$0.3} \textcolor{teal}{(+0.35)}} & \textbf{95.10\scalebox{0.8}{$\pm$0.1} \textcolor{teal}{(+0.13)}} & \textbf{94.93\scalebox{0.8}{$\pm$0.1} \textcolor{teal}{(+0.18)}} & \textbf{65.39\scalebox{0.8}{$\pm$0.2} \textcolor{teal}{(+0.03)}} & \textbf{65.90\scalebox{0.8}{$\pm$0.0} \textcolor{teal}{(+1.38)}} & \textbf{65.38\scalebox{0.8}{$\pm$0.2} \textcolor{teal}{(+1.27)}}\\
\midrule
FlexMatch & 94.63\scalebox{0.8}{$\pm$0.1} \textcolor{red}{(-0.01)} & 94.86\scalebox{0.8}{$\pm$0.3} \textcolor{teal}{(+0.08)} & 93.82\scalebox{0.8}{$\pm$0.7} \textcolor{red}{(-0.23)} & 66.16\scalebox{0.8}{$\pm$0.2} \textcolor{teal}{(+0.16)} & 64.57\scalebox{0.8}{$\pm$0.1} \textcolor{red}{(-1.01)} & 64.22\scalebox{0.8}{$\pm$0.4} \textcolor{red}{(-0.45)}\\
FlexMatch (real only) & 94.64\scalebox{0.8}{$\pm$0.1} & 94.78\scalebox{0.8}{$\pm$0.2} & 94.05\scalebox{0.8}{$\pm$0.2} & 66.00\scalebox{0.8}{$\pm$0.2} & 65.58\scalebox{0.8}{$\pm$0.1} & 64.67\scalebox{0.8}{$\pm$0.1} \\
\textbf{Ours-FlexMatch} & \textbf{95.32\scalebox{0.8}{$\pm$0.1} \textcolor{teal}{(+0.68)}} & \textbf{94.92\scalebox{0.8}{$\pm$0.0} \textcolor{teal}{(+0.14)}} & \textbf{95.12\scalebox{0.8}{$\pm$0.3} \textcolor{teal}{(+1.07)}} & \textbf{66.61\scalebox{0.8}{$\pm$0.2} \textcolor{teal}{(+0.61)}} & \textbf{66.36\scalebox{0.8}{$\pm$0.1} \textcolor{teal}{(+0.78)}} & \textbf{66.40\scalebox{0.8}{$\pm$0.3} \textcolor{teal}{(+1.73)}} \\
\midrule
SoftMatch & 94.29\scalebox{0.8}{$\pm$0.1} \textcolor{red}{(-0.69)} & 93.87\scalebox{0.8}{$\pm$0.1} \textcolor{red}{(-1.14)} & 93.63\scalebox{0.8}{$\pm$0.2} \textcolor{red}{(-1.12)} & 66.41\scalebox{0.8}{$\pm$0.2} \textcolor{red}{(-0.10)} & 66.60\scalebox{0.8}{$\pm$0.1} \textcolor{teal}{(+0.32)} & 65.53\scalebox{0.8}{$\pm$0.3} \textcolor{red}{(-0.27)} \\
SoftMatch (real only) & 94.98\scalebox{0.8}{$\pm$0.1} & 95.01\scalebox{0.8}{$\pm$0.2} & 94.75\scalebox{0.8}{$\pm$0.1} & 66.51\scalebox{0.8}{$\pm$0.2} & 66.28\scalebox{0.8}{$\pm$0.1} & 65.80\scalebox{0.8}{$\pm$0.3} \\
\textbf{Ours-SoftMatch} & \textbf{95.05\scalebox{0.8}{$\pm$0.1} \textcolor{teal}{(+0.07)}} & \textbf{95.24\scalebox{0.8}{$\pm$0.1} \textcolor{teal}{(+0.23)}} & \textbf{94.80\scalebox{0.8}{$\pm$0.3} \textcolor{teal}{(+0.05)}} & \textbf{67.52\scalebox{0.8}{$\pm$0.2} \textcolor{teal}{(+1.01)}} & \textbf{68.03\scalebox{0.8}{$\pm$0.3} \textcolor{teal}{(+1.75)}} & \textbf{67.31\scalebox{0.8}{$\pm$0.3} \textcolor{teal}{(+1.51)}} \\
\bottomrule
\end{tabular}
}
\end{center}
\caption{Classification accuracy (\%) with various synthetic ratios $\alpha$ on CIFAR-10 and CIFAR-100 with 25 labels per class.}
\label{tab1}
\end{table*}
\begin{table}[t]
\centering
\begin{center}
\resizebox{\linewidth}{!}{
\begin{tabular}{l|cc|cc}
\toprule
Dataset & \multicolumn{2}{c|}{CIFAR-10} & \multicolumn{2}{c}{CIFAR-100}  \\
\midrule
\# of labeled data & 25 & 50 & 25 & 50 \\
\midrule
FixMatch &  94.80\scalebox{0.8}{$\pm$0.2} & 94.84\scalebox{0.8}{$\pm$0.1} & 64.14\scalebox{0.8}{$\pm$0.0} & 68.69\scalebox{0.8}{$\pm$0.1} \\
FixMatch (real only) &  94.97\scalebox{0.8}{$\pm$0.1} & 94.92\scalebox{0.8}{$\pm$0.1} & 64.52\scalebox{0.8}{$\pm$0.1} & 69.35\scalebox{0.8}{$\pm$0.2} \\
\textbf{Ours-FixMatch} & \textbf{95.10\scalebox{0.8}{$\pm$0.1}} & \textbf{95.25\scalebox{0.8}{$\pm$0.1}} & \textbf{65.90\scalebox{0.8}{$\pm$0.0}} & \textbf{70.13\scalebox{0.8}{$\pm$0.1}} \\
\midrule
FlexMatch &  94.86\scalebox{0.8}{$\pm$0.3} & 95.14\scalebox{0.8}{$\pm$0.1}  & 64.57\scalebox{0.8}{$\pm$0.1}  & 69.12\scalebox{0.8}{$\pm$0.1}  \\
FlexMatch (real only) &  94.78\scalebox{0.8}{$\pm$0.2} & 94.92\scalebox{0.8}{$\pm$0.0} & 65.58\scalebox{0.8}{$\pm$0.1} & 69.35\scalebox{0.8}{$\pm$0.0} \\
\textbf{Ours-FlexMatch} &  \textbf{94.92\scalebox{0.8}{$\pm$0.0}} & \textbf{95.61\scalebox{0.8}{$\pm$0.0}} & \textbf{66.36\scalebox{0.8}{$\pm$0.1}} & \textbf{70.12\scalebox{0.8}{$\pm$0.1}} \\
\midrule
SoftMatch &  93.87\scalebox{0.8}{$\pm$0.1} & 95.17\scalebox{0.8}{$\pm$0.2} & 66.60\scalebox{0.8}{$\pm$0.1} & 69.68\scalebox{0.8}{$\pm$0.1} \\
SoftMatch (real only) &  95.01\scalebox{0.8}{$\pm$0.2} & 95.08\scalebox{0.8}{$\pm$0.1} & 66.28\scalebox{0.8}{$\pm$0.1} & 70.01\scalebox{0.8}{$\pm$0.1} \\
\textbf{Ours-SoftMatch} &  \textbf{95.24\scalebox{0.8}{$\pm$0.1}} & \textbf{95.45\scalebox{0.8}{$\pm$0.1}} & \textbf{68.03\scalebox{0.8}{$\pm$0.3}} & \textbf{70.33\scalebox{0.8}{$\pm$0.0}} \\
\bottomrule
\end{tabular}
}
\end{center}
\caption{Classification accuracy (\%) with various number of labeled data per class on CIFAR-10 and CIFAR-100 with synthetic ratio $\alpha=0.5$.}
\label{tab2}
\end{table}

\noindent\textbf{Baseline methods.} We evaluate and compare RSMatch with the above-mentioned SSL methods: FixMatch~\cite{sohn2020fixmatch}, FlexMatch~\cite{zhang2021flexmatch}, and SoftMatch~\cite{chen2023softmatch}. Moreover, we compare with some robust SSL methods to investigate if they can tackle the RS-SSL task. We select OpenMatch~\cite{saito2021openmatch} and IOMatch~\cite{li2023iomatch}, which are designed for the open-set problem, and CAFA~\cite{huang2021universal}, which is designed for both open-set and domain difference problems.

\subsection{Experimental results}

\textbf{SSL methods on RS-SSL benchmark.} We evaluate three representative SSL methods on our RS-SSL benchmark using CIFAR-10 and CIFAR-100. Table~\ref{tab1} shows the results on various synthetic ratios $\alpha$. Table~\ref{tab2} shows the results on various numbers of labeled data.  \textbf{Real only} denotes we manually remove the synthetic images in the unlabeled data. Thus, it can be referenced as the ideal scenario in which the SSL methods are not affected by any synthetic images. We can conclude from the results that \textbf{SSL methods struggled to utilize existing unlabeled synthetic images.} For instance, FixMatch experiences a 1.89\% drop in classification accuracy when an equal number of synthetic images are added alongside the original real images on CIFAR-10. However, advancements in SSL methods have not addressed the RS-SSL task, as both FlexMatch and SoftMatch also struggle with this issue.

\noindent\textbf{The effectiveness of RSMatch.} On the contrary, the results show that our RSMatch can further utilize the unlabeled synthetic data to improve the SSL performance. For example, RSMatch improves by 1.75\% in accuracy on CIFAR-100 with an additional 50\% synthetic data using SoftMatch. The same conclusion is drawn in Table~\ref{tab2}, where we fix the synthetic ratio and select various numbers of labeled images. For example, RSMatch improves by 0.78\% on CIFAR-100 with 50 labeled images per class using FixMatch.

\begin{table}[t]
\centering
\begin{center}
\resizebox{\linewidth}{!}{
\begin{tabular}{l|ccc|c}
\toprule
Dataset & \multicolumn{3}{c|}{TinyImageNet} & ImageNet  \\
\midrule
Synthetic ratio ($\alpha$) & 0.3 & 0.5 & 1.0 & 0.5 \\
\midrule
FixMatch &  39.52\scalebox{0.8}{$\pm$0.2} & 38.40\scalebox{0.8}{$\pm$0.2} & 36.72\scalebox{0.8}{$\pm$0.3} & 64.85\scalebox{0.8}{$\pm$0.1} \\
FixMatch (real only) & 40.26\scalebox{0.8}{$\pm$0.3} & 39.30\scalebox{0.8}{$\pm$0.3} & 38.40\scalebox{0.8}{$\pm$0.3} & 65.29\scalebox{0.8}{$\pm$0.3} \\
\textbf{Ours-FixMatch} &  \textbf{40.49\scalebox{0.8}{$\pm$0.3}} & \textbf{40.60\scalebox{0.8}{$\pm$0.3}} & \textbf{40.21\scalebox{0.8}{$\pm$0.1}} & \textbf{65.94\scalebox{0.8}{$\pm$0.4}} \\
\bottomrule
\end{tabular}
}
\end{center}
\caption{Classification accuracy (\%) with various synthetic ratio $\alpha$ on TinyImageNet and ImageNet with 10\% labels per class.}
\label{tab3}
\end{table}
\begin{table}[t]
\centering
\begin{center}
\begin{tabular}{l|ccc}
\toprule
Dataset & \multicolumn{3}{c}{CIFAR-10}  \\
\midrule
Synthetic ratio ($\alpha$) & 0.3 & 0.5 & 1.0 \\
\midrule
FixMatch &  95.10 & 94.75 & 92.88 \\
OpenMatch &  69.84 & 72.03 & 69.34 \\
IOMatch &  94.39 & 94.77 & 94.41 \\
CAFA &  94.91 & 94.81 & 94.35 \\
\midrule
\textbf{Ours-FixMatch} & \textbf{95.29} & \textbf{95.08} & \textbf{94.92} \\
\bottomrule
\end{tabular}

\end{center}
\caption{Comparision with robust SSL methods with various synthetic ratios on CIFAR-10 with 25 labels per class.}
\label{tab4}
\end{table}
\begin{table}[t]
\centering
\begin{center}
\begin{tabular}{c|c|c|cc}
\toprule
\makecell{Ind.\\models} & \makecell{Class-wise\\CSQueue} & \makecell{Dummy\\head} & \makecell{Deepfake\\Acc.} & \makecell{SSL\\Acc.} \\
\midrule
 \ding{55} & \ding{55} & \ding{55} & - & 94.75 \\
\midrule
\ding{55} &  &  & 65.12 & 92.68 \\
 & \ding{55} &  & 65.48 & 92.46 \\
 &  & \ding{55} & 95.55 & 94.91 \\
 \midrule
\multicolumn{3}{c|}{Ours-FixMatch} & 95.11 & 95.08 \\
\bottomrule
\end{tabular}
\end{center}
\caption{Ablation studies of each component in RSMatch on CIFAR-10 with 25 labels per class and the synthetic ratio $\alpha=0.5$.}
\label{tab5}
\end{table}

\noindent\textbf{Experiments on large-scale datasets.} We further evaluate RSMatch on large-scale TinyImageNet and ImageNet with more classes, samples, and higher resolution. The results in Table~\ref{tab3} show that the synthetic data also have a negative impact on these challenging datasets. However, RSMatch can still utilize synthetic data to improve SSL performance. For example, RSMatch improves by 1.81\% on TinyImageNet with $\alpha=1.0$.

\begin{figure}
\centering
\includegraphics[width=0.99\linewidth]{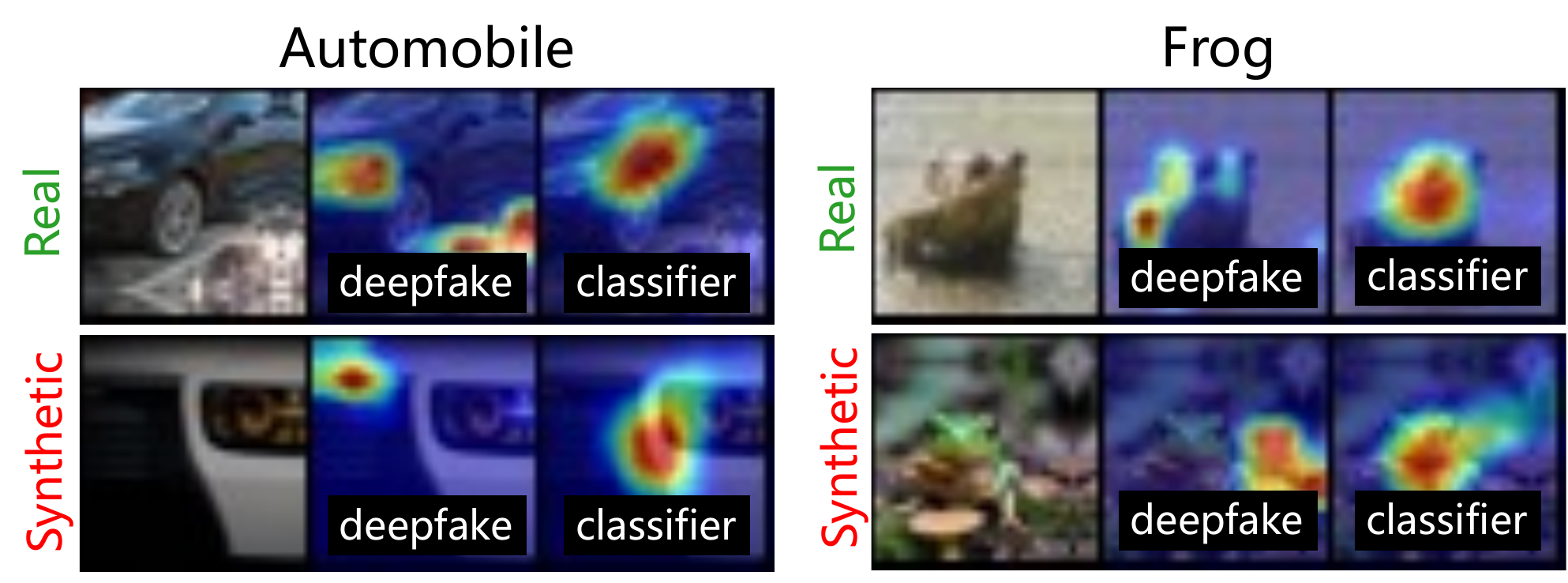}
\caption{The GradCAM++ visualization of the deepfake detector and the classifier on real and synthetic images of difference classes from CIFAR-10. Note that these two models have different attention regions. Thus prove the effectiveness of using two individual models.}
\label{fig6}
\end{figure}

\noindent\textbf{Evaluation of robust SSL methods.} Despite conventional SSL methods, there are also robust SSL methods focusing on the problems of unlabeled data like open-set and domain difference. We evaluate several robust SSL methods on the RS-SSL benchmark in Table~\ref{tab4}. Note that CAFA~\cite{huang2021universal} is designed for both open-set and domain differences in the unlabeled data. However, as we've mentioned, their setting is all the unlabeled data are from another specific domain. Thus, their domain-matching strategy, treating all unlabeled data as another domain, failed on the RS-SSL task. The results further prove the necessity of proposing the RS-SSL task and RSMatch.

\noindent\textbf{Ablation studies on RSMatch components.} We conduct ablation studies on each component of RSMatch in Table~\ref{tab5}. Row~1 denotes our baseline FixMatch. Row~2 shows that the absence of an individual deepfake detector (i.e., \textbf{ind. models}) impacts identification accuracy. Meanwhile, as shown in the GradCAM++~\cite{chattopadhay2018grad} visualization in Fig.~\ref{fig6}, the deepfake detector and classifier have different focuses on the same image, thus proving the necessity of using two individual networks. Row~3 shows that the single queue without class-wise structure results in worse synthetic data identification accuracy. As we mentioned, the single queue will be dominated by specific classes. Row~4 shows that the synthetic data cannot be utilized for training to improve SSL performance without the dummy head. The results prove the effectiveness of each component in RSMatch. 

\begin{figure}
\centering
\includegraphics[width=0.99\linewidth]{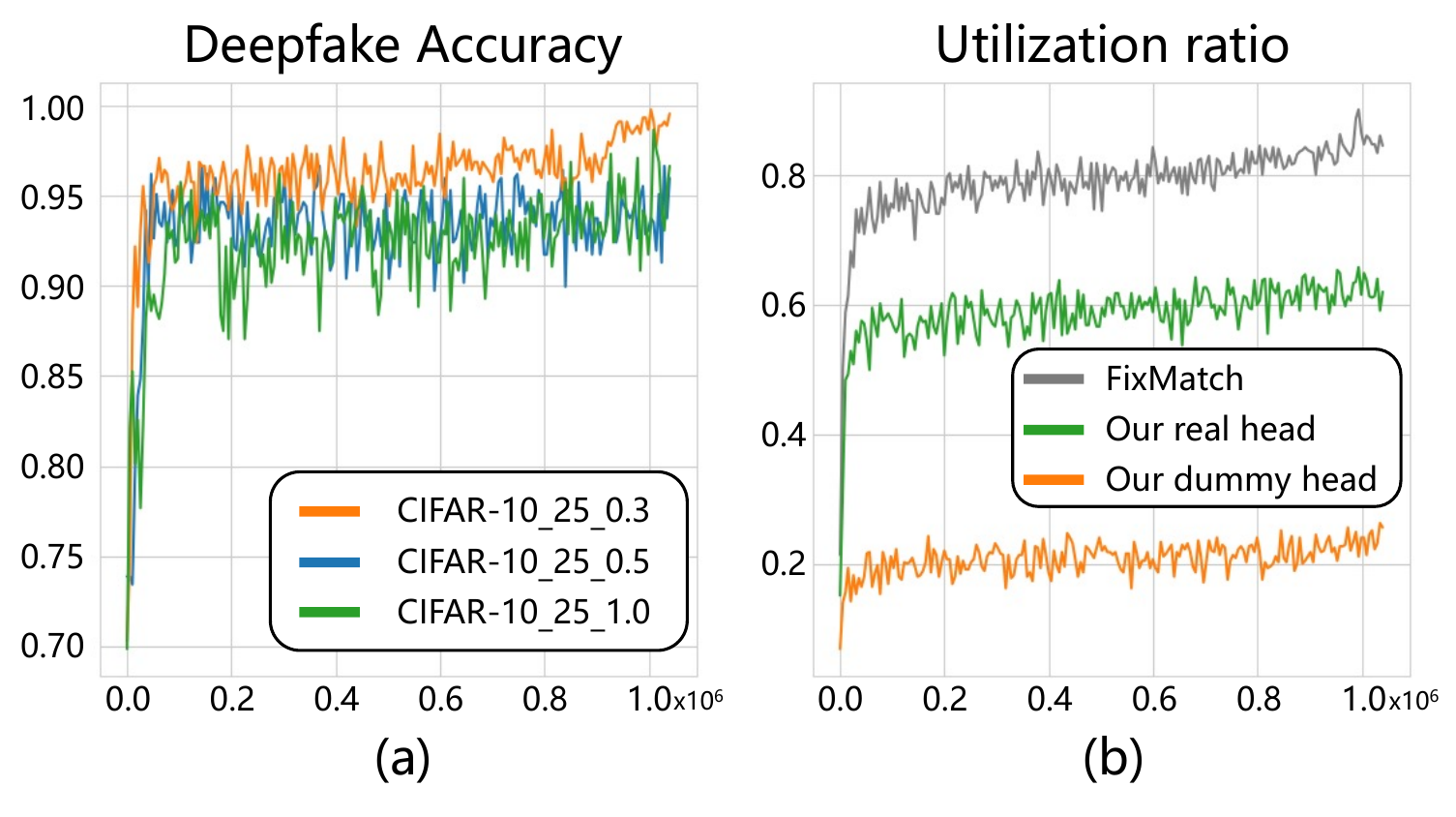}
\caption{\textbf{(a)} The deepfake detection accuracy curve on unlabeled data during training with various synthetic ratios on CIFAR-10 with 25 labeled images per class. \textbf{(b)} The utilization ratio of unlabeled samples during training (samples with confidence score over the SSL threshold) of FixMatch and the two heads in our method on CIFAR-10 with 25 labeled images per class and the synthetic ratio of 0.5.}
\label{fig7}
\end{figure}

\noindent\textbf{Analysis of SSL process.} In Fig.~\ref{fig7}, we further analyze the SSL process from the accuracy curve of the deepfake detector and the utilization ratio of unlabeled data (\ie, unlabeled data with confidence score over the SSL threshold) during training. Fig.~\ref{fig7}~(a) shows that our proposed deepfake detector and CSQueue can successfully distinguish real and synthetic images \textbf{under the condition that no synthetic image labels are provided} in the RS-SSL task. Fig.~\ref{fig7}~(b) shows that FixMatch will directly utilize a large amount of synthetic data during training, thus resulting in low performance. Meanwhile, RSMatch can divide these synthetic data into the dummy head, thus eliminating the negative impact and improving the performance.

\section{Conclusion}
In this paper, we propose and address a new challenging task, Real-Synthetic Hybrid SSL (RS-SSL), to investigate the impact of the increasingly generated and uploaded synthetic data on SSL. We first set up the RS-SSL benchmark to evaluate the performance of existing SSL methods. We found the current SSL methods are not ready to face the challenge of including existing synthetic images. Then, we propose RSMatch, a new SSL method that can identify the synthetic images in the unlabeled data without any provided synthetic labels and further utilize them to improve SSL performance. RSMatch can effectively reduce the need to specifically care for synthetic data when collecting unlabeled data from public image sources such as the Internet. Therefore, our research further advances the practicality of semi-supervised learning.
{
    \small
    \bibliographystyle{ieeenat_fullname}
    \bibliography{main}
}


\end{document}